\RequirePackage{fix-cm}
\documentclass[smallextended]{svjour3}
\usepackage{amsmath,amsfonts,graphicx,float,color,enumerate,multirow, longtable, csvsimple,ifthen,dsfont,amssymb, mathtools, url}
\usepackage{epstopdf}
\usepackage{subcaption}
\usepackage{tikz,pgfplots}
\usetikzlibrary{trees}
\usepackage{pgf-umlsd}
\usepackage{algorithm, algpseudocode}
\pgfplotsset{compat=1.8}
\usetikzlibrary{shapes,arrows}
\usepackage{cases}
\usepackage{booktabs}

\usepackage{tikz}
\usetikzlibrary{arrows,positioning, calc}
\tikzstyle{sprout}=[draw,fill=yellow!15,circle,minimum size=20pt,inner sep=2pt]
\tikzstyle{leaf}=[draw,fill=black!15,circle,minimum size=20pt,inner sep=2pt, rectangle]

\newcommand{\reals}{\mathbb{R}}
\newcommand{\vq}{\boldsymbol q}

\newcommand{\vx}{\boldsymbol x}

\newcommand{\vi}{\boldsymbol i}

\newcommand{\D}{{\rm d}}

\newcommand{\cross}[3]{$\mathbin{\draw[#3, line width=.2ex] (-0.25+#1,-0.3125+#2) -- (0.25+#1,0.3125+#2) (-0.5+#1,0.0+#2) -- (0.5+#1,0.0+#2) ;}$}

\definecolor{green_figures}{RGB}{0,153,0}
\definecolor{red_figures}{RGB}{255,25,25}

\begin{document}

\title{An Explainable Bayesian Decision Tree Algorithm}

\author{Giuseppe Nuti  \and \mbox{Llu\'is Antoni Jim\'enez Rugama}
 \and Andreea-Ingrid Cross}

\institute {Giuseppe Nuti, \at UBS,  Europa-Strasse 1, 8152 Opfikon, Switzerland, \email{giuseppe.nuti@ubs.com}
		\and 
		Llu\'is Antoni Jim\'enez Rugama, \at UBS, 1285 6th Ave, New York, NY 10019, US, \email{lluis.jimenez-rugama@ubs.com}
		\and 
		Andreea-Ingrid Cross, \at UBS, 5 Broadgate Cir, London EC2M 2QS, UK, \email{andreea-ingrid.cross@ubs.com}
}

\date{Received: date / Accepted: date}

\maketitle

\begin{abstract}
	Bayesian Decision Trees are known for their probabilistic interpretability. However, their construction can sometimes be costly. In this article we present a general Bayesian Decision Tree algorithm applicable to both regression and classification problems. The algorithm does not apply Markov Chain Monte Carlo and does not require a pruning step. While it is possible to construct a weighted probability tree space we find that one particular tree, the greedy-modal tree (GMT), explains most of the information contained in the numerical examples. This approach performs similarly to Random Forests on various benchmark data sets; notably, the Bayesian tree achieves similar accuracy utilizing a single (relatively shallow) tree, allowing for a fully explainable technique which is often a prerequisite for applications in finance or the medical industry.
\end{abstract}

\section{Introduction}

	Decision trees are popular machine learning techniques applied to both classification and regression tasks. This technique is characterized by the resulting model, which is encoded as a tree structure. All nodes in a tree can be observed and understood, thus decision trees are considered white boxes. In addition, the tree structure can return the output with considerably fewer computations than other, more complex, machine learning techniques. Some examples of classical decision tree algorithms include the CART \cite{Brei84} and the C4.5 \cite{Qui93}. These algorithms have later been improved in \cite{Frie00}, boosted trees, and extended to several trees in \cite{Brei01}, Random Forests.
	 
	 The first probabilistic approaches, also known as Bayesian Decision Trees, were introduced in \cite{Bun92}, \cite{Chip98}, and \cite{Den98}. The first article proposed a deterministic algorithm while the other two are based on Markov Chain Monte Carlo convergence. The main challenge is that the space of all possible tree structures is large, and as noted in \cite{Chic96}, \cite{Chic97}, the search for an optimal decision tree given a scoring function is NP-hard.
	 
	 In this article, we propose an algorithm similar to \cite{Bun92} where we explicitly model the entire \emph{tree generation process} as opposed to only providing a probabilistic score of the possible partitions. This allows us to view the pruning aspect probabilistically instead of relying on a heuristic algorithm.

	We start in Section \ref{sec:overview} with an overview of the article's Bayesian trees. Section \ref{sec:partition_space} provides our building blocks, the partition probability space, followed by the trees probability space construction in Section \ref{sec:trees_space}. We present some numerical results in Section \ref{sec:numerical_examples} showing that the greedy-modal Bayesian Decision Tree works well for various publicly available data sets. Even though we focus on the classification task as part of this article's evaluation section, this algorithm can equivalently be applied to a regression task. Finally, some conclusive remarks follow in Section \ref{sec:discussion}.

	\section{Bayesian Trees Overview}\label{sec:overview}
	We define $\mathcal{D}=\{(\vx_i,y_i)\}_{i=1}^n$ a data set of $n$ independent observations. Points $\vx=(x^1,\dots,x^d)$ in $\reals^d$ describe the features\footnote{When our data set contains non-ordinal categorical features, we extend the feature space with as many features as the number of categories minus one, and assign these features the values 0 or 1, where 1 is assigned to the column the category belongs to -- akin to a dummy variable approach. For instance, a data set $\{((4,\text{car}), 0), ((0,\text{house}), 1), ((3,\text{house}), 1), ((1,\text{dream}), 0),((2,\text{car}), 0)\}$ would be transformed into $\{((4,1,0,0), 0), ((0,0,1,0), 1), ((3,0,1,0), 1), ((1,0,0,1), 0), ((2,1,0,0), 0)\}$.} of each observation whose outcome $y$ is randomly sampled from $Y_{\vx}$. The distribution of $Y_{\vx}$ will determine the type of problem we are solving: a discrete random variable translates into a classification problem whereas a continuous random variable translates into a regression problem. The beta function will be specially useful to compute the likelihood for the classification examples in this article,
\begin{equation}\label{eq:beta_fun}
B\left(z^1,\dots,z^C\right)=\frac{\prod_{c=1}^C\Gamma(z^c)}{\Gamma\left(\sum_{c=1}^C z^c\right)},
\end{equation}
$C$ different classes and $\Gamma$ the gamma function
	
	The data set $\mathcal{D}$ is sampled from a data generation process. We divide this process into two steps: first, a point $\vx$ is sampled in $\reals^d$; second, the outcome $Y_{\vx}$ is sampled given $\vx$. In this article we do not consider any prior knowledge of the generation of locations $\vx$. Hence, we will focus on the distribution of $Y_{\vx}$.
This conditional distribution is assumed to be encoded under a tree structure created following a set of simple recursive rules based on $\{\vx_i\}_{i=1}^n$, namely the \emph{tree generation process}: starting at the root, we determine if we are to expand the current node with two leaves under a predefined probability. This probability may depend on the current depth. If there is no expansion, the process terminates for this node, otherwise we choose which dimension within d is the split applicable to. Once the dimension is chosen, we assume that the specific location of the split across the available range is distributed across all \emph{distinct} points within that range. After we have determined the specific location of the split, we assume that the process continues iteratively by again determining if each new leaf is to be split again. When there are no more nodes with a split or each leaf contains only one distinct set of observations, the tree is finalized.

\begin{figure}[!h]
    \centering
    \begin{subfigure}[c]{1.0\textwidth}
	\centering
		\begin{tikzpicture}
		\fill[green_figures] (0.0,0.0) circle (3.0pt);
		\fill[green_figures] (0.5,0.0) circle (3.0pt);
		\fill[green_figures] (1.25,0.0) circle (3.0pt);
		\fill[red_figures] (1.5,0.0) circle (3.0pt);
		\fill[red_figures] (1.75,0.0) circle (3.0pt);
	\end{tikzpicture}
	\caption{$B(4,3)/B(1,1)$}
    \label{fig:onedimsplit0}
\end{subfigure}
~
\begin{subfigure}[c]{1.0\textwidth}
	\centering
		\begin{tikzpicture}
		\fill[green_figures] (0.0,0.0) circle (3.0pt);
		\fill[green_figures] (0.5,0.0) circle (3.0pt);
		\fill[green_figures] (1.25,0.0) circle (3.0pt);
		\fill[red_figures] (1.5,0.0) circle (3.0pt);
		\fill[red_figures] (1.75,0.0) circle (3.0pt);
		\draw[black](0.25,-0.25) -- (0.25,0.25);
	\end{tikzpicture}
	\caption{$B(2,1)B(3,3)/B(1,1)^2$}
    \label{fig:onedimsplit1}
\end{subfigure}
~
\begin{subfigure}[c]{1.0\textwidth}
	\centering
	\begin{tikzpicture}
		\fill[green_figures] (0.0,0.0) circle (3.0pt);
		\fill[green_figures] (0.5,0.0) circle (3.0pt);
		\fill[green_figures] (1.25,0.0) circle (3.0pt);
		\fill[red_figures] (1.5,0.0) circle (3.0pt);
		\fill[red_figures] (1.75,0.0) circle (3.0pt);
		\draw[black](0.875,-0.25) -- (0.875,0.25);
	\end{tikzpicture}
	\caption{$B(3,1)B(2,3)/B(1,1)^2$}
    \label{fig:onedimsplit2}
\end{subfigure}
~
\begin{subfigure}[c]{1.0\textwidth}
	\centering
	\begin{tikzpicture}
		\fill[green_figures] (0.0,0.0) circle (3.0pt);
		\fill[green_figures] (0.5,0.0) circle (3.0pt);
		\fill[green_figures] (1.25,0.0) circle (3.0pt);
		\fill[red_figures] (1.5,0.0) circle (3.0pt);
		\fill[red_figures] (1.75,0.0) circle (3.0pt);
		\draw[black](1.375,-0.25) -- (1.375,0.25);
	\end{tikzpicture}
	\caption{$B(4,1)B(1,3)/B(1,1)^2$}
    \label{fig:onedimsplit3}
\end{subfigure}
~
\begin{subfigure}[c]{1.0\textwidth}
	\centering
	\begin{tikzpicture}
		\fill[green_figures] (0.0,0.0) circle (3.0pt);
		\fill[green_figures] (0.5,0.0) circle (3.0pt);
		\fill[green_figures] (1.25,0.0) circle (3.0pt);
		\fill[red_figures] (1.5,0.0) circle (3.0pt);
		\fill[red_figures] (1.75,0.0) circle (3.0pt);
		\draw[black](1.625,-0.25) -- (1.625,0.25);
	\end{tikzpicture}
	\caption{$B(4,2)B(1,2)/B(1,1)^2$}
    \label{fig:onedimsplit4}
\end{subfigure}
\caption{Likelihoods $L(\mathcal{D}|\Pi)$ of our data set $\{(0,1), (0.5,1), (1.25, 1), (1.5,0),$ $(1.75,0)\}$ given some partitions of $\reals$, outcomes following a Bernoulli distribution, and assuming a prior Beta$(1,1)$.}\label{fig:onedimexpartitions}
\end{figure}
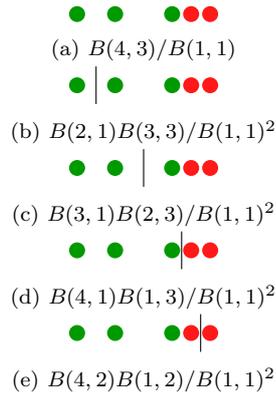

	Given the generating process above, it is clear that the building block of the Bayesian Decision Trees is to consider partitions of $\reals^d$ that better explain the outcomes under a probabilistic approach. All points in the same set of a partition $\Pi$ share the same outcome distribution, i.e. $Y$ does not depend on $\vx$. Hence, assuming a prior for the distribution parameters of $Y$ we can obtain the likelihood of each set in a partition. Since all observations are assumed independent, the total partition likelihood $L(\mathcal{D}|\Pi)$ is obtained by multiplying the likelihoods of each set. Figure \ref{fig:onedimexpartitions} shows some partition likelihoods for points $\{0, 0.5, 1.25, 1.5, 1.75\}$ and categorical outcomes $\{1, 1, 1, 0, 0\}$. From the likelihoods in Figure \ref{fig:onedimexpartitions}, we see that the highest likely partition is \eqref{fig:onedimsplit3}, i.e. $\{(-\infty,1.375], (1.375,\infty)\}$. We do not imply however, that the other partitions cannot occur.
	
	Finally, given an additional prior on the partition space, we can obtain the probability of each partition given $\mathcal{D}$, i.e. $\mathbb{P}(\Pi|\mathcal{D})=L(\mathcal{D}|\Pi)\mathbb{P}(\Pi)/\mathbb{P(\mathcal{D})}$. While $\mathbb{P}(\Pi|\mathcal{D})$ is the probability of a partition, the information contained in each partition set is the posterior of the $Y$ parameters. In Figure \ref{fig:onedimexpartitions}, the posterior distributions are Beta$(4, 1)$ if $x\leq 1.375$, and Beta$(1, 3)$ otherwise. This partition space is described in detail in Section \ref{sec:partition_space}.
	
	Given a non-trivial partition, we can create additional partition sub-spaces for each partition set. We can iterate the sub-partitioning until the trivial partition is chosen or the set has only one observation. Thus, a specific choice of partitions with sub-partitions can be represented as a tree. We include the trivial partition in each probability space to train the trees without the need of a pruning step. Section \ref{sec:trees_space} constructs the probability space of binary trees, i.e. those for which each non-trivial partition splits the data into exactly two sets. 

	\section{Partition Probability Space}\label{sec:partition_space}
	In this section we define the partition space which is the building block for the Bayesian Decision Trees. Let $\mathcal{D}=\{(\vx_i,y_i)\}_{i=1}^n$ be a data set with $n$ independent observations in $M\times\reals$, $M\subseteq\reals^d$. A partition $\Pi=\{M_1,\dots,M_k\}$ of $M$ divides $M$ into disjoint subsets $M_1,\dots,M_k$ such that $M = \cup M_w$. As a consequence, the data set $\mathcal{D}$ will be split into $\mathcal{D}_{M_1},\dots,\mathcal{D}_{M_k}$, where the observation $(\vx_i,y_i)$ belongs to $\mathcal{D}_{M_w}$ if and only if $\vx_i\in M_w$. All  $Y_{M_w}$ sampled within region $M_w$ follow the same distribution with probability measure $\rho_{\vq}$ and parameters $\vq$ in $\reals^{s}$.

	We assume a prior distribution $\gamma_{M_w}$ for parameters $\vq$. Then, the likelihood of our data given the prior is,
	\begin{equation}\label{eq:likelihood}
	L(\mathcal{D}_{M_w}) = \begin{cases}
	\int_{\reals^{s}}\left(\prod_{y_i\in\mathcal{D}_{M_w}}\rho_{\vq}(y_i)\right)\gamma_{M_w}(\vq)\D \vq, & \mathcal{D}_{M_w}\neq \emptyset, \\
	1,  & \mathcal{D}_{M_w}= \emptyset,
	\end{cases}	
	\end{equation}
	where $y_i\in\mathcal{D}_{M_w}$ is the set of all $y_i$ such that $(\vx_i,y_i)$ is in $\mathcal{D}_{M_w}$. The likelihood of our data given the partition is,
	\begin{equation}\label{eq:total_likelihood}
	L({\mathcal{D}}|\Pi) = \prod_{w=1}^kL(\mathcal{D}_{M_w}).
	\end{equation}	
	In addition, if we provide a prior probability measure for partitions, $p(\Pi)$ over $\Omega_{\Pi}$, the updated probability of a partition given our data is,
	\begin{equation}\label{eq:prob_part_norm}
		p(\Pi|\mathcal{D}) = \frac{L(\mathcal{D}|\Pi)p(\Pi)}{\mathbb{P}(\mathcal{D})} = \frac{L(\mathcal{D}|\Pi)p(\Pi)}{\int_{\Omega_\Pi}L(\mathcal{D}|\Pi)p(\Pi)\D \Pi}.
	\end{equation}
For practical reasons we will work with the non-normalized probabilities,
	\begin{equation}\label{eq:prob_part}
	\widetilde{p}(\Pi|\mathcal{D}) = L(\mathcal{D}|\Pi)p(\Pi)\propto p(\Pi|\mathcal{D}).
	\end{equation}
Finally, the posterior distribution of $\vq$ in each region $M_w$ is
\[
\mu(\vq)=\frac{L(y_i\in\mathcal{D}_{M_w}|\vq)\gamma_{M_w}(\vq)}{\int_{\reals^{s}} L(y_i\in\mathcal{D}_{M_w}|\vq)\gamma_{M_w}(\vq)\D \vq}.
\]

	There are uncountably many possible partitions of $M$. To construct binary trees, we reduce the space of partitions $\Omega_{\pi}$ to the trivial partition $\Pi_0=\{M\}$, and partitions of the form $\Pi_{r,m}=\{\{\vx\in M \text{ such that } x^r\leq h_m\},\{\vx\in M \text{ such that } x^r> h_m\}\}$ for some real $h_m$ and some dimension $r=1,\dots,d$. This choice of partitions will make the algorithm invariant to feature dimension scaling. Furthermore, we note that any partition splitting two neighboring observations will result in the same posteriors and likelihood of the data. For instance, all partitions of the form $\{(-\infty,h], (h,\infty)\}$, $1.25<h<1.5$ in Figure \ref{fig:onedimexpartitions} are equivalent. We borrow the following idea from Support Vector Machines (SVM) \cite{Cort95}: from all partitions splitting two distinct neighboring observations, we only consider the maximum margin classifier, i.e. the partition that splits at their mid-point. Hence, the partition space $\Omega_{\Pi}$ is finite by definition. In addition, with this $\Omega_{\pi}$ we can aggregate all probabilities of partitions along dimension $r$, i.e. $p(r|\mathcal{D})=\sum_m p(\Pi_{r,m}|\mathcal{D})$, to evaluate the importance of each dimension. Figure \ref{fig:onedimexpartitions} shows all possible partitions for the example in Section \ref{sec:overview}.
	
\begin{figure}
	\begin{subfigure}{.5\textwidth}
		\centering
		\includegraphics[width=1.0\textwidth]{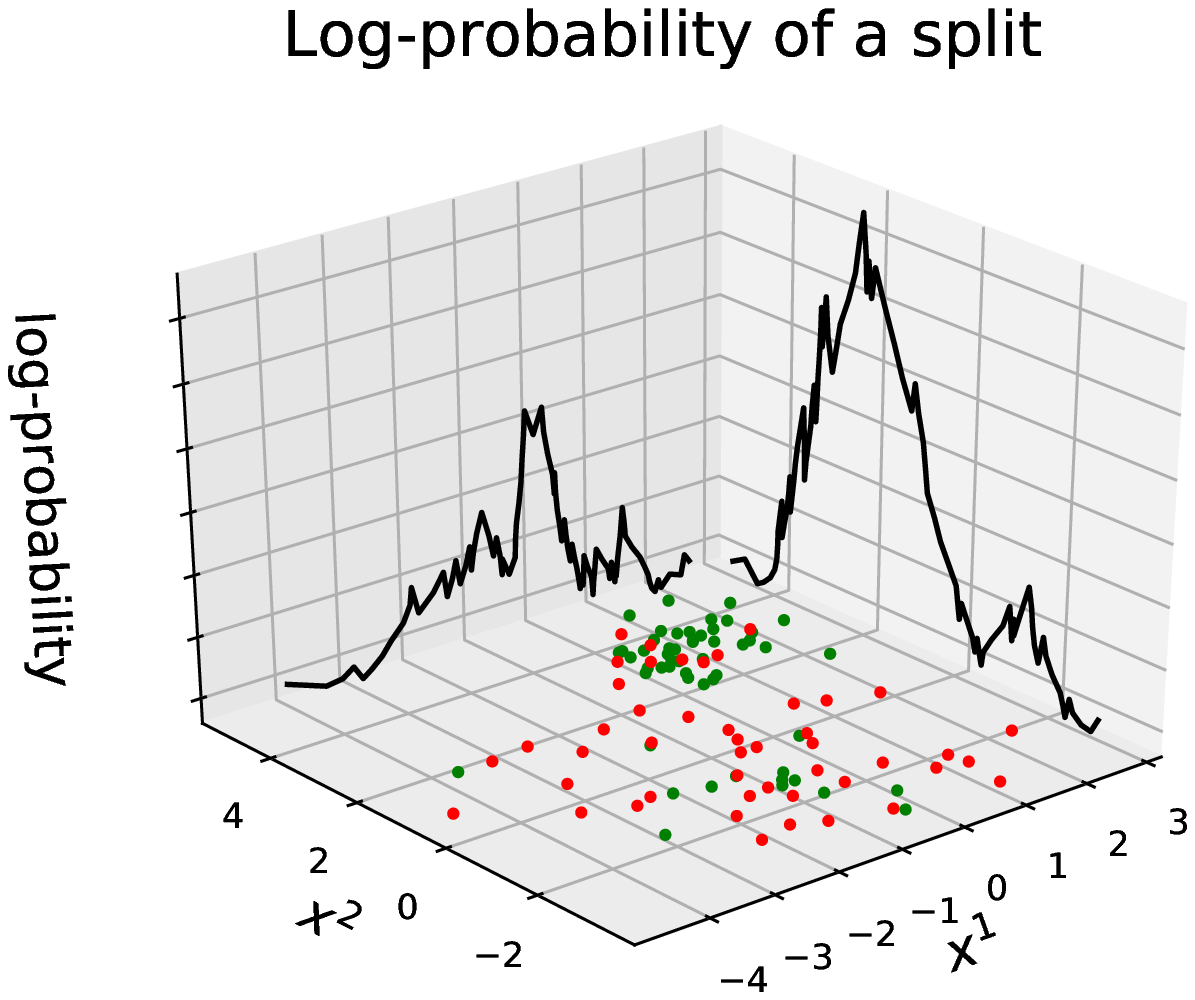}
		\caption{}\label{fig:logprobabilities}
	\end{subfigure}
	\begin{subfigure}{.5\textwidth}
		\includegraphics[width=1.0\textwidth]{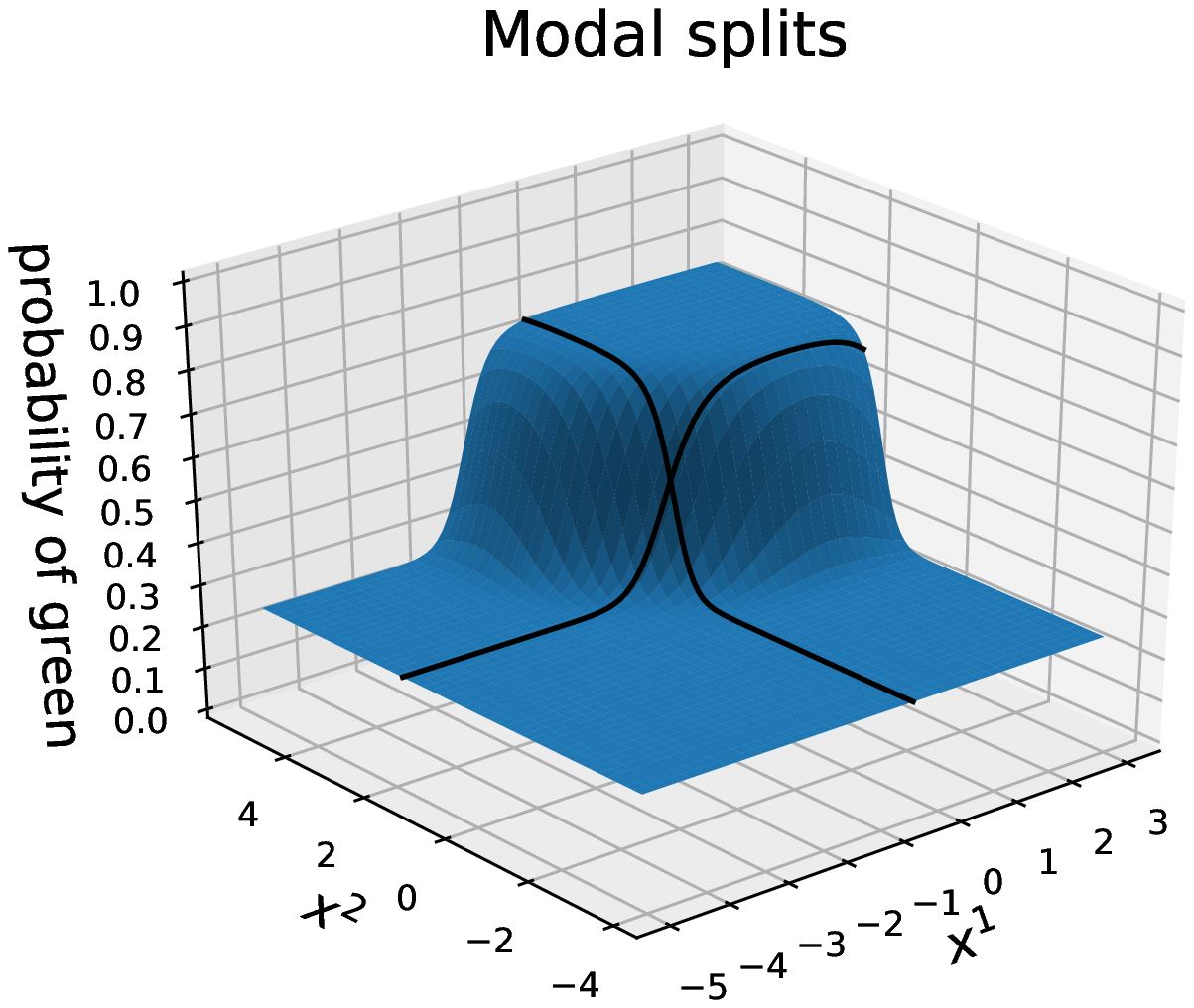}
		\caption{}\label{fig:modalsplits}
	\end{subfigure}
	\caption{Example of a data set whose outcomes are either green or red. The location of the points is sampled from a mixture of two Gaussian distributions with equal probability. One distribution draws outcomes from a Bernoulli distribution with probability 0.25, while the other with probability 0.75. Figure \ref{fig:logprobabilities}: log-probabilities of all possible non-trivial partitions given the data set. Figure \ref{fig:modalsplits}: actual probability of a data point being green in addition to the modal split of each dimension.}\label{fig:partition_space}
	\end{figure}

	As an example, Figure \ref{fig:partition_space} provides the log-probabilities of the partition space for a data set in $\reals^2$ whose outcomes are drawn from Bernoulli random variables. The samples are generated from two distributions equally likely, i.e. we draw from each distribution with probability 0.5. The first distribution is a multivariate Gaussian with mean $(-1,-1)$ and covariance $2\,\rm I$. Points sampled from this distribution have a probability of 0.25 of being green. The second distribution is another multivariate Gaussian with mean $(1,3)$ and covariance $\rm 0.5I$. In this case, the probability of a sample of being green is 0.75. Because the means of these Gaussian distributions are further apart along the $x^2$ axis, the highest probable partitions are found in this dimension.
	
	One natural choice of partition is the mode. In the Appendices, we provide Algorithm \ref{alg:findNodePointer} which returns the modal partition. For the special case of the classification problem, we also provide Algorithm \ref{alg:findNodePointerCat}. To improve stability and efficiency, both algorithms work with the log-probabilities and assume we know the sorted indices for our points, namely $\vi_1,\dots,\vi_n$ such that $x_{i_j^r}^r \leq x_{i_{j+1}^r}^r$ for all $r=1,\dots,d$ and $j=1,\dots,n-1$. In addition, Algorithm \ref{alg:findNodePointerCat} assumes $Y$ follows a multivariate Bernoulli distribution with outcomes in $\{1,\dots,C\}$, and the prior $\gamma$ belongs to the Dirichlet family. The prior $\gamma$ will be characterized by its parameters $(\alpha^1,\dots,\alpha^C)$. 

	\section{Bayesian Decision Trees}\label{sec:trees_space}
\begin{figure}
\centering
\begin{tikzpicture}[very thick,level/.style={sibling distance=45mm/#1}]
\node [sprout] (r){\small Raining?}
  child {
    node [sprout] {\small Cloudy?}
    child {
      node [leaf] {10\%}
      edge from parent [->] node [left] {no}
    }
    child {
      node [leaf] {40\%}
      edge from parent [->] node [right] {yes}
    }
    edge from parent [->] node [left] {no}
  }
  child {
    node [sprout] {\small $0.1$ in/h?}
    child {
      node [leaf] {50\%}
      edge from parent [->] node [left] {less}
    }
    child {
      node [sprout] {\small 5 h?}
      child {
      	node [leaf] {90\%}
      	edge from parent [->] node [left] {less}
      }
      child {
      	node [leaf] {70\%}
      	edge from parent [->] node [right] {more}
      }
      edge from parent [->] node [right] {more}
    }
    edge from parent [->] node [right] {yes}
  };
\end{tikzpicture}
\caption{Example of a binary tree. Leaves are marked in grey and sprouts in yellow. This tree provides the probability of raining in one hour.}\label{fig:rain_tree}
\end{figure}
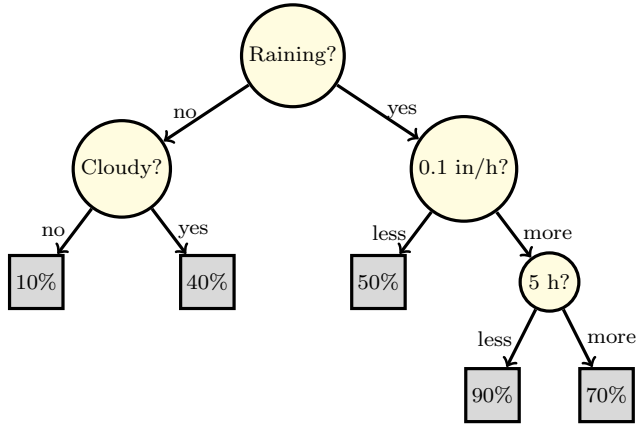

	Trees are directed acyclic graphs formed of nodes with a single root, and where every two connected nodes are connected by a unique path. We classify the nodes as either \emph{leaves} or \emph{sprouts}. While leaves are terminal nodes containing the model information, sprouts point to additional child nodes. If the number of child nodes is always two, we call the tree a \emph{binary} tree. Each sprout contains a question or rule whose answer will lead to one of its children. Starting from the root, which is the only node without a parent, we follow the tree path until we reach a leaf. Figure \ref{fig:rain_tree} shows an example of a binary tree.
	
	The partition space from Section \ref{sec:partition_space} is the preamble to constructing the Bayesian trees. Each partition from $\Omega_\Pi$ can be identified to a tree node. If $\Pi$ is the trivial partition $\Pi_0$, the node becomes a \emph{leaf}, otherwise a \emph{sprout}. By construction, non-trivial partitions will only have two subsets of $M$: the lower subset $M_L=\{\vx\in M \text{ such that } x^r\leq h\}$, and the upper subset $M_U=\{\vx\in M \text{ such that } x^r > h\}$. If we choose a non-trivial partition $\Pi$, we can construct additional partition spaces on $M_L$ given $\mathcal{D}_{M_L}$, and $M_U$ given $\mathcal{D}_{M_U}$. We can repeat this process until we choose all trivial partitions, i.e. leaves, as per equation \eqref{eq:prob_tree}. The leaves will contain the posterior distributions $\mu(\vq)$, as shown in Figure \ref{fig:tree_example}, and the not normalized probability of the tree will be,
\begin{equation}\label{eq:prob_tree}
f(M,\mathcal{D}) = 
\begin{cases}
f(M_L,\mathcal{D}_{M_L})f(M_U,\mathcal{D}_{M_U}), \quad & \text{$\Pi$ is not trivial}, \\
\widetilde{p}(\Pi|\mathcal{D}), \quad & \text{$\Pi$ is trivial}.
\end{cases}
\end{equation}
This probability will be normalized over the set of trees we construct. Note that when the data set contains less than two distinct observations, $\Omega_\Pi=\{\Pi_0\}$.
	
\begin{figure}
\begin{subfigure}{.5\textwidth}
\centering
\begin{tikzpicture}[x=7.5pt,y=7.5pt,yscale=1,xscale=1]
	\draw[fill={gray},fill opacity=0.5] (0,1.25) -- (4,6.25) -- (14,6.25) -- (10,1.25) -- cycle;
	\draw[->,color=black,thick] (-1.4,0.75) -- (3.6,0.75);
	\draw[->,color=black,thick] (-1.4,0.75) -- (0.6,3.25);
	\draw ++(1,0) node {\small $x^1$};
	\draw ++(-1.25,2.5) node {\small $x^2$};
	\foreach \i in {1,...,9}
	{
		\draw[thin ,color=black!75] (0+\i,1.25) -- (4+\i,6.25);
		\draw[thin ,color=black!75] (0+\i*0.4,1.25+\i*0.5) -- (10+\i*0.4,1.25+\i*0.5);
	}

	\draw[fill={gray}  ,fill opacity=0.5] (-1,-5) -- (3,0) -- (8,0) -- (4,-5) -- cycle;
	\foreach \i in {1,...,5}
	{
		\draw[thin ,color=black!75] (-1+\i,-5) -- (3+\i,0);
		\draw[thin ,color=black!75] (-1+\i*0.4,-5+\i*0.5)  -- (4+\i*0.4,-5+\i*0.5);
		\draw[thin ,color=black!75] (0.6+\i*0.4,-3+\i*0.5)  -- (5.6+\i*0.4,-3+\i*0.5);
	}

	\draw[fill={gray}  ,fill opacity=0.5] (6,-5) -- (10,0) -- (15,0) -- (11,-5) -- cycle;
	\foreach \i in {1,...,5}
	{
		\draw[thin ,color=black!75] (6+\i,-5) -- (10+\i,0);
		\draw[thin ,color=black!75] (6+\i*0.4,-5+\i*0.5)  -- (11+\i*0.4,-5+\i*0.5);
		\draw[thin ,color=black!75] (7.6+\i*0.4,-3+\i*0.5)  -- (12.6+\i*0.4,-3+\i*0.5);
	}

	\draw[fill={gray}  ,fill opacity=0.5] (5.6,-9.75) -- (6.6,-8.5) -- (11.6,-8.5) -- (10.6,-9.75) -- cycle;
	\draw[fill={gray}  ,fill opacity=0.5] (7.4,-7.5) --  (10.4,-3.75) -- (15.4,-3.75) -- (12.4,-7.5) -- cycle;
	\foreach \i in {1,...,4}
	{
		\draw[thin ,color=black!75] (5.6+\i,-9.75) -- (6.6+\i,-8.5);
		\draw[thin ,color=black!75] (7.4+\i,-7.5) -- (10.4+\i,-3.75);
	}
	\foreach \i in {1,2,5,6,7,8,9,10,11}
	{
		\draw[thin ,color=black!75] (5.6+\i*0.4,-9.75+\i*0.5)  -- (10.6+\i*0.4,-9.75+\i*0.5);
	}
	
	\cross{5}{2.75}{red_figures}
	\cross{4}{4.75}{red_figures}
	\cross{1.5}{2.0}{red_figures}
	\cross{6}{5.0}{red_figures}
	\cross{9.0}{4.0}{red_figures}
	\cross{10.5}{3.25}{red_figures}
	\cross{7.5}{1.75}{green_figures}
	\cross{9.0}{2.0}{green_figures}
	
	\cross{5-1}{1.75-5.25}{red_figures}
	\cross{4-1}{3.75-5.25}{red_figures}
	\cross{1.5-1}{1.0-5.25}{red_figures}
	\cross{6-1}{4.0-5.25}{red_figures}
	\cross{9.0+1}{3.0-5.25}{red_figures}
	\cross{10.5+1}{2.25-5.25}{red_figures}
	\cross{7.5+1}{0.75-5.25}{green_figures}
	\cross{9.0+1}{1.0-5.25}{green_figures}
	
	\cross{9.0+1+0.4}{3.0-4.25-5.25+0.5}{red_figures}
	\cross{10.5+1+0.4}{2.25-5.25-4.25+0.5}{red_figures}
	\cross{7.5+1-0.4}{0.75-5.25-4.25-0.5}{green_figures}
	\cross{9.0+1-0.4}{1.0-5.25-4.25-0.5}{green_figures}
\end{tikzpicture} \caption{}\label{fig:tree_layers}
\end{subfigure}
\begin{subfigure}{.5\textwidth}
\centering
\begin{tikzpicture}[very thick,level/.style={sibling distance=30mm/#1}]
\node [sprout] {$x^1$?}
  child {
    node [leaf] {\scriptsize Beta$(1,5)$}
    edge from parent [->] node [left] {\scriptsize $x^1\leq 5$}
  }
  child {
    node [sprout] {$x^2$?}
    child {
      node [leaf] {\scriptsize Beta$(3,1)$}
      edge from parent [->] node [left] {\scriptsize $x^2\leq 2.5$}
    }
    child {
      node [leaf] {\scriptsize Beta$(1,3)$}
      edge from parent [->] node [right] {\scriptsize $x^2> 2.5$}
    }
    edge from parent [->] node [right] {\scriptsize $x^1> 5$}
  };
\end{tikzpicture} \caption{}\label{fig:tree}
\end{subfigure}
\caption{Example of a tree for a 2-categories data set. Figure \ref{fig:tree_layers}: the data set is displayed three times. The first layer corresponds to the data set before any split. The second layer displays the lower and upper sets resulting from splitting along dimension $x$. The third layer is an additional split of the upper subset along dimension $y$. Figure \ref{fig:tree}: resulting tree with posterior distributions for $\vq$ assuming that outcomes are generated from Bernoulli random variables and the conjugate prior is Beta$(1,1)$.}\label{fig:tree_example}
\end{figure}

	We can construct a tree and obtain $f(M,\mathcal{D})$ at the same time, see Algorithm \ref{alg:fill_tree} in the Appendices. Within this algorithm, the method \break \verb'choose_partition' is not specified and should contain the search logic to choose the children partitions. The cost of Algorithm \ref{alg:fill_tree} is $\mathcal{O}(\text{cost}(n)\ln(n))$ where $\text{cost}(n)$ is the cost of \verb'choose_partition'. The main problem is to construct the trees whose probability is highest and that are structurally different. To start with, a particularly interesting Bayesian Decision Tree is the one obtained by choosing the modal partition at each step. We will call it the \emph{greedy-modal tree} (GMT). This tree can be constructed by specifying \verb'choose_partition' to be \verb'find_modal_partition' from Algorithms \ref{alg:findNodePointer} or \ref{alg:findNodePointerCat}. If we choose Algorithm \ref{alg:findNodePointerCat} to be the \verb'choose_partition', the average cost of Algorithm \ref{alg:fill_tree} becomes $\mathcal{O}(dn\ln(n))$. If we want to add more trees that are highly likely but structurally different, we can construct them as we do for the \emph{greedy-modal tree} but by choosing different roots.
	
	Algorithms \ref{alg:findNodePointer} and \ref{alg:findNodePointerCat} only look at one level ahead. As suggested in \cite{Bun92}, we could optimize the construction by looking at several levels ahead at the expense of increasing the order of $\text{cost}(n)$. However, Section \ref{sec:numerical_examples} shows that the GMT constructed with Algorithm \ref{alg:findNodePointerCat} performs well in practice.
	
	When we query for a point in the tree space, the answer is the posterior $\mu(\vq)$. Nonetheless, we can also return an expected value. In the classification problems from Section \ref{sec:numerical_examples}, we return the tree weighted average of $\mathbb{E}[\vq]$.

	\section{Numerical Examples}\label{sec:numerical_examples}
	In this Section we apply Algorithms \ref{alg:findNodePointerCat} and \ref{alg:fill_tree} to construct the GMT. We assume that the outcomes, 0 or 1, are drawn from Bernoulli random variables. The prior distribution is set to Beta$(10,10)$ and each tree will return the expected probability of drawing the outcome $0$, namely $\mathbb{E}[\vq]$. The prior probabilities for each partition will be $p(\Pi)=1-0.99^{1+{\ell}}$ if $\Pi$ is the trivial partition, and $p(\Pi)=0.99^{1+{\ell}}/dN$ otherwise, where $N$ is the number of non-trivial partitions along the dimension, and $\ell\in\mathbb{N}_0$ is the depth at which the partition space lies. Note that dividing the non-trivial partition prior probability by $d$ is implicitly assuming a uniform prior distribution on the dimension space. One could also include in the analysis the posterior distributions of each dimension to visualize which features are most informative. As an alternative to this prior, we could use the partition distance weighted approach from \cite{Bun92}. These are the default settings which we apply to all of the data-sets studied here.
	
	The accuracy 	is measured as a percentage of correct predictions. Each prediction will simply be the maximum probability outcome. If there is a tie, we choose category 0 by default. We compare the results to Decision Trees (DT) \cite{Brei84}, and Random Forests (RF) \cite{Brei01} computed with the \verb|DecisionTreeClassifier| and \verb|RandomForestRegressor| objects from the Python scikit-learn package \cite{scikit-learn}. For reproducibility purposes, we set all random seeds to 0. In the case of RF, we enable bootstrapping to improve its performance and fix the number of trees to 5. The GMT results are generated with Java although we also provide the Python module with integration into scikit-learn in \cite{GMT19}.

		\subsection{UCI Data Sets}
		We test the GMT on some data sets from the University of California, Irvine (UCI) database \cite{Dua17}. We compute the accuracy the DT, RF, and GMT. Except for the Ripley set where a test set with 1000 points is provided, we apply a shuffled 10-fold cross validation. Results are shown in Table \ref{table:accuracy} and training time in Table \ref{table:timing}.
		
\begin{table}[h]
\center
\begin{tabular}{c|c|c|c|c|c|c|}
\cline{2-6}  & \multicolumn{5}{c|}{Accuracy} \\ \cline{2-7}
\multicolumn{1}{c|}{} & \multicolumn{1}{c|}{$d$} & $n$ & DT & RF & GMT & GMT - RF \\ \hline
\multicolumn{1}{|l|}{Heart}  & \multicolumn{1}{c|}{20}   & 270   & 76.3\% & 78.5\% & 83.0\% & 4.5\% \\ \hline
\multicolumn{1}{|l|}{Credit} & \multicolumn{1}{c|}{23}   & 30\,000 & 72.6\% & 78.1\% & 82.0\% & 3.9\%  \\ \hline
\multicolumn{1}{|l|}{Haberman} & \multicolumn{1}{c|}{3} & 306   & 65.0\% &68.3\% & 71.9\% & 3.6\% \\ \hline
\multicolumn{1}{|l|}{Seismic} & \multicolumn{1}{c|}{18}   & 2\,584  & 87.7\% & 91.5\% & 93.2\% & 1.7\%   \\ \hline
\multicolumn{1}{|l|}{Ripley} & \multicolumn{1}{c|}{2} & 250/1\,000 & 83.8\% & 87.9\% & 87.6\% & -0.3\% \\ \hline
\multicolumn{1}{|l|}{Gamma} & \multicolumn{1}{c|}{10}   & 19\,020 & 81.4\% & 85.6\% & 85.2\% & -0.4\% \\ \hline
\multicolumn{1}{|l|}{Diabetic} & \multicolumn{1}{c|}{19}   & 1\,151  & 62.6\%& 64.8\% & 63.5\% & -1.3\%  \\ \hline
\multicolumn{1}{|l|}{EEG} & \multicolumn{1}{c|}{14}   & 14\,980 & 84.0\%& 88.6\% & 81.2\% & -7.4\%  \\ \hline
\end{tabular}
\caption{Accuracy of DT, RF, and GMT for several data sets. Except for the Ripley data set, we apply a 10-fold cross validation to each test. Results are sorted by relative performance, starting from hightest accuracy difference between GMT and RF.}\label{table:accuracy}
\end{table}

	The results reveal some interesting properties of GMT. Noticeably, the tree constructed in GMT seems to perform well in general with a training time between the DT and RF times. In all cases, the DT accuracy is lower than the RF accuracy. The cases in which RF outperforms the GMT are the Ripley, Gamma, Diabetic and EEG data sets. The accuracy difference between GMT and RF may show that the assumptions for both models are different. Furthermore, the fixed prior Beta$(10,10)$ might not be optimal for some of the data sets. The EEG data set is the least performing set for GMT compared to both, DT and RF. One reason may be that some information is hidden in lower levels, i.e. information that cannot be extracted by selecting the modal partition at each level.
	
\begin{table}[h]
\center
%\begin{tabular}{l|r|r|r|r|r|r|}
%\cline{2-7}
% & \multicolumn{3}{c|}{Train time (ms)} & \multicolumn{3}{c|}{Query time ($\mu$s)} \\ \cline{2-7}
%\multicolumn{1}{c|}{} & \multicolumn{1}{c|}{DT} & \multicolumn{1}{c|}{RF} & \multicolumn{1}{c|}{GMT} & \multicolumn{1}{c|}{DT} & \multicolumn{1}{c|}{RF} & \multicolumn{1}{c|}{GMT} \\ \hline
%\multicolumn{1}{|l|}{Heart} & 0.0 & 4.7 & 4.7  & 0.0 & 57.8 & 0.0  \\ \hline
%\multicolumn{1}{|l|}{Credit} & 521.0 & 1388.4 & 837.7 & 0.5 & 0.0 & 0.0  \\ \hline
%\multicolumn{1}{|l|}{Haberman} & 0.0 & 4.7 & 1.6  & 0.0 & 51.0 & 0.0  \\ \hline
%\multicolumn{1}{|l|}{Ripley} & 0.0 & 1.6 & 0.0  & 0.0 & 0.0 & 0.0  \\ \hline
%\multicolumn{1}{|l|}{Seismic} & 10.9 & 28.1 & 20.3 & 6.0 & 0.0 & 0.0  \\ \hline
%\multicolumn{1}{|l|}{Gamma} & 254.3 & 634.9 & 238.7 & 0.0 & 0.8 & 2.4  \\ \hline
%\multicolumn{1}{|l|}{Diabetic} & 7.8 & 31.2 & 10.9 & 0.0 & 0.0 & 0.0  \\ \hline
%\multicolumn{1}{|l|}{EEG} & 115.4 & 338.5 & 263.6 & 0.0 & 0.0 & 0.0  \\ \hline
%\end{tabular}
\begin{tabular}{l|r|r|r|}
\cline{2-4}
 & \multicolumn{3}{c|}{Train time (ms)} \\ \cline{2-4}
\multicolumn{1}{c|}{} & \multicolumn{1}{c|}{DT} & \multicolumn{1}{c|}{RF} & \multicolumn{1}{c|}{GMT} \\ \hline
\multicolumn{1}{|l|}{Heart} & 0.0 & 4.7 & 3.8 \\ \hline
\multicolumn{1}{|l|}{Credit} & 521.0 & 1388.4 & 823.2 \\ \hline
\multicolumn{1}{|l|}{Haberman} & 0.0 & 4.7 & 0.6 \\ \hline
\multicolumn{1}{|l|}{Seismic} & 10.9 & 28.1 & 18.9  \\ \hline
\multicolumn{1}{|l|}{Ripley} & 0.0 & 1.6 & 0.0   \\ \hline
\multicolumn{1}{|l|}{Gamma} & 254.3 & 634.9 & 252.1 \\ \hline
\multicolumn{1}{|l|}{Diabetic} & 7.8 & 31.2 & 10.6  \\ \hline
\multicolumn{1}{|l|}{EEG} & 115.4 & 338.5 & 271.4 \\ \hline
\end{tabular}
\caption{Training time in milliseconds per fold.}\label{table:timing}
\end{table}

	\subsection{Smoothing}
\begin{figure}[h]
		\centering
		\includegraphics[width=0.5\textwidth]{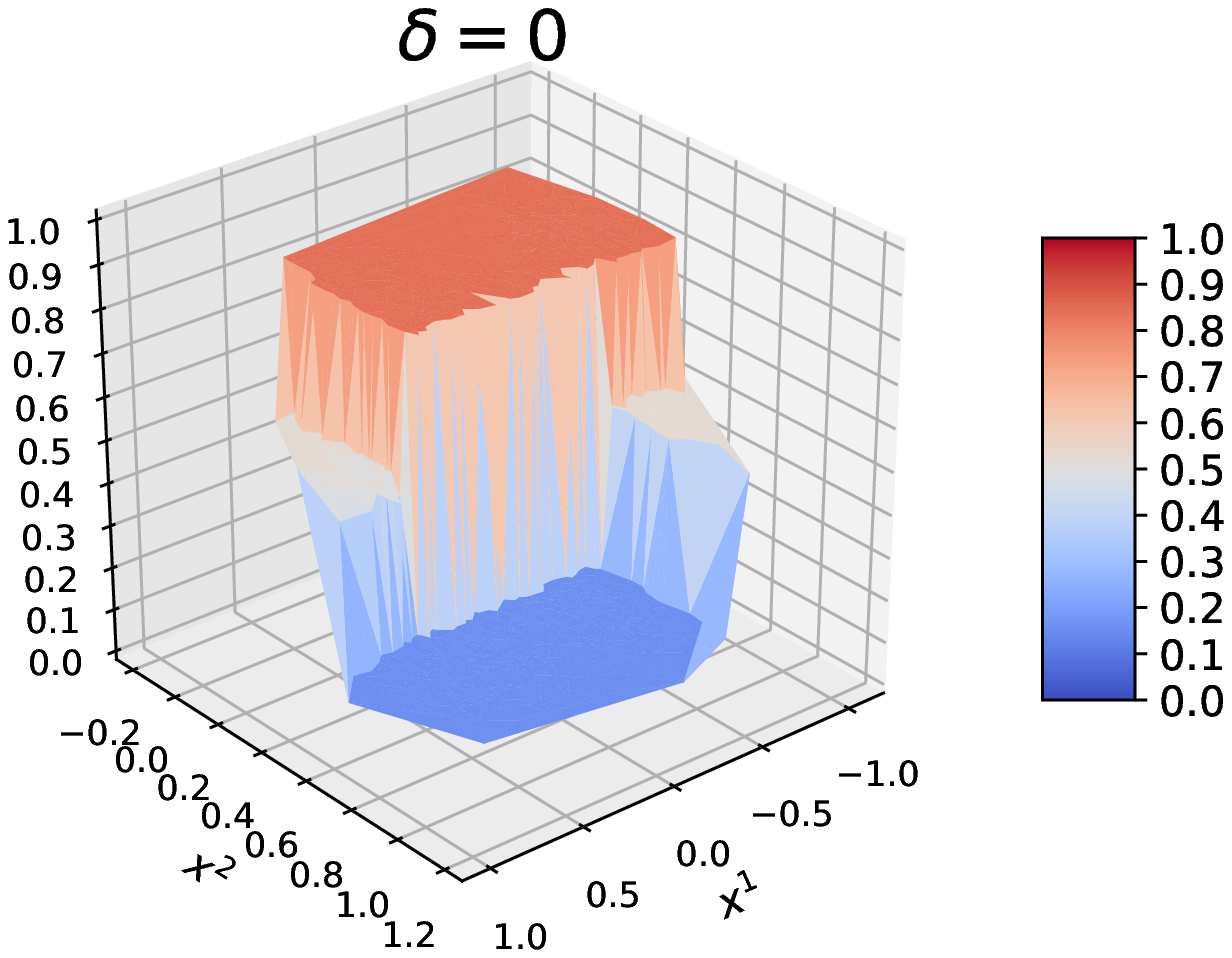}\includegraphics[width=0.5\textwidth]{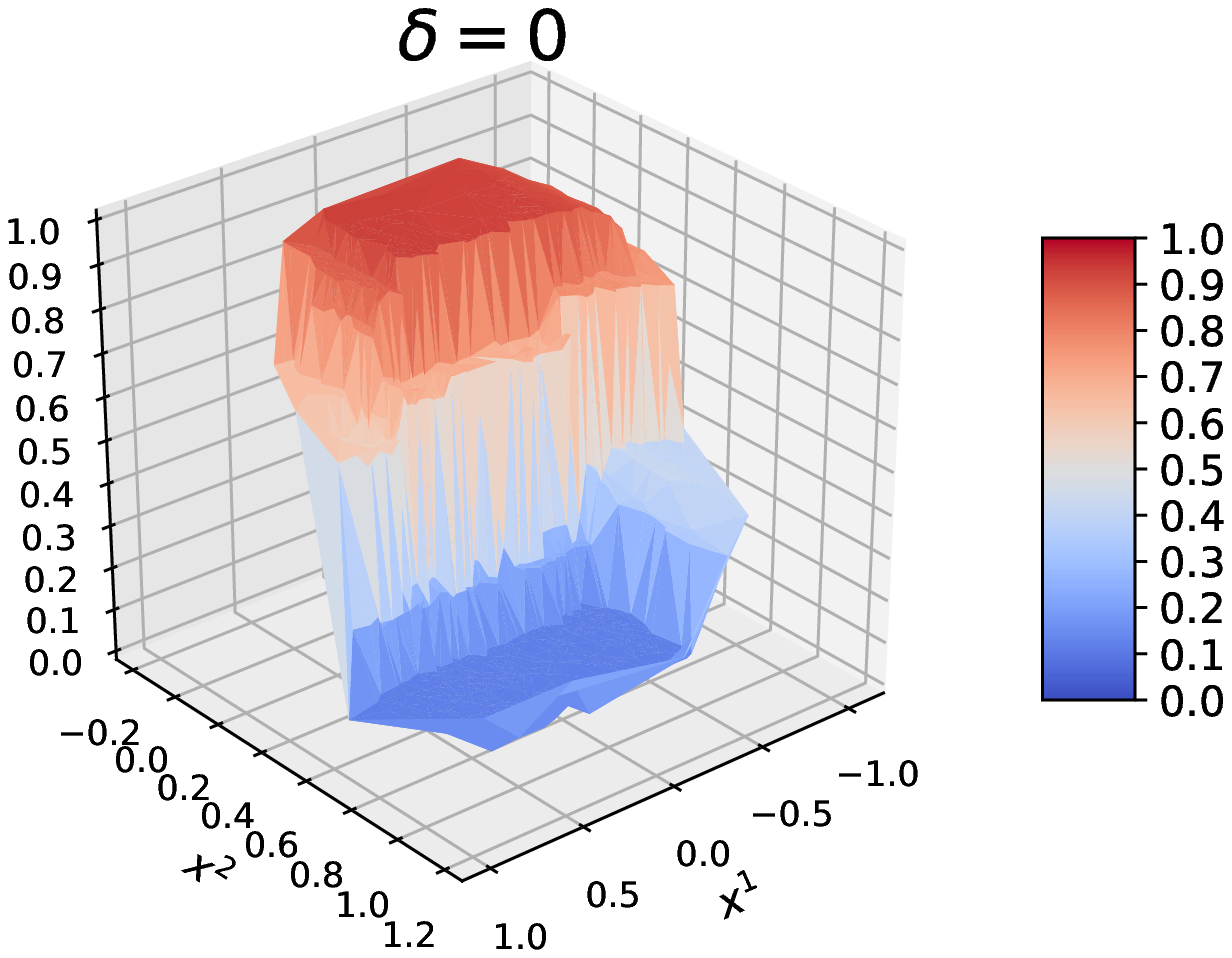}
		\caption{Probabilities of outcome 0 for the Ripley test set with different $\delta$. To allow for a deeper exploration, we fix $p(\Pi)=0.99/2N$ for non-trivial partitions. On the left $\delta=0$, and on the right $\delta=0.1$.}\label{fig:ripley_smooth}
	\end{figure}

	In some cases, the nature of the problem warrants some smoothness in the solution, i.e. we do not desire abrupt changes in the posterior distributions for small regions. Intuitively, when we zoom in a region, we expect the parameters $\vq$ of $Y$ to be more certain. Hence, we can reduce the variance of the prior distribution at each split.  In particular, for the examples in this Section, we propose to modify the prior distribution as follows: let Beta$(\alpha_0, \alpha_1)$ be the prior distribution at one partition space. For a specific non-trivial partition in this space, define $n_0,n_1$ the $\mathcal{D}_L$ total number of samples of each category, and $m_0,m_1$ the $\mathcal{D}_U$ total number of samples of each category. The prior distributions that we further use in $\mathcal{D}_L$ and $\mathcal{D}_U$ are Beta$(\alpha_0+\delta n_0, \alpha_1+\delta n_1)$ and Beta$(\alpha_0+\delta m_0, \alpha_1+\delta m_1)$ respectively, where $\delta$ is a proportion of the total number of samples. If we choose $\delta=0$, we return to the original formulation with constant prior. Figure \ref{fig:ripley_smooth} displays the effects of the smoothing on GMT.
	
	\section{Discussion and Future Work}\label{sec:discussion}
	The proposed GMT is a Bayesian Decision Tree that reduces the training time by avoiding any Markov Chain Monte Carlo sampling or the pruning step. The GMT numerical example results have similar predictive power to RF. This approach may be most useful where the ability to explain the model is a requirement as it has the same interpretability as the DT but a performance similar to RF. Hence, the advantages of the GMT are that it can be easily understood and takes less time than RF to train. Furthermore, the ability to specify a prior probability may be particularly suitable to some problems. It still remains to find a more efficient way to explore meaningful trees and improve performance.
	
	As an extension, we would like to assess the performance of this algorithm on regression problems and experiment with larger partition spaces such as the SVM hyperplanes. Another computational advantage not explored is parallelization, which would allow for a more exhaustive exploration of the tree probability space. Finally, the smoothing concept has been briefly introduced: despite its intuitive appeal, it still remains to explore the theoretical foundations behind it.
	
%\section*{References}
\bibliographystyle{plain}

\newpage
\appendix
\section*{Appendix}

\begin{algorithm}
\caption{Find the modal partition in $\Omega_{\Pi}$ for the general problem.}
\label{alg:findNodePointer}
\begin{algorithmic}[1]
\Procedure{find\_modal\_partition}{$\mathcal{D}, \gamma$}
   	\State $loglike \gets \text{get\_loglike}(\mathcal{D},\gamma)$\Comment{Applies equation \eqref{eq:likelihood}.}
   	\State $logprob \gets loglike+\ln(p(\Pi_{0}))$\Comment{Computes $\ln(\widetilde{p})$.}
   	\State $\Pi^*\gets \text{create\_trivial\_part}(loglike, logprob)$
    \For {each dimension $r$ in $1,\dots,d$}
	    \State $m \gets 1$ \Comment{Identifies each non-trivial partition.}
    	\For {each observation $j$ in $1,\dots,n-1$}
        	\If {$x^r_{i_j^r} \neq x^r_{i_{j+1}^r}$}
        		\State $c \gets (x^r_{i_j^r} + x^r_{i_{j+1}^r})/2$
        		\State $loglike_L \gets \text{get\_loglike}\left(\{\vx_{i_a^r},y_{i_a^r}\}_{a=1}^j,\gamma\right)$\Comment{Applies equation \eqref{eq:likelihood}.}
				\State $loglike_U \gets \text{get\_loglike}\left(\{\vx_{i_a^r},y_{i_a^r}\}_{a=j+1}^n,\gamma\right)$\Comment{Applies equation \eqref{eq:likelihood}.}
				\State $loglike \gets loglike_L + loglike_U$\Comment{Applies eq \eqref{eq:total_likelihood}.}
			   	\State $logprob \gets loglike+\ln(p(\Pi_{r,m}))$\Comment{Computes $\ln(\widetilde{p})$.}
        		\State $m \gets m + 1$
        		\If {$\text{get\_logprob}(\Pi^*)<logprob$}
        			\State $\Pi^*\gets \text{create\_part}(c,r,loglike, logprob)$
        		\EndIf
        	\EndIf
		\EndFor
	\EndFor
	\State Return $\Pi^*$
\EndProcedure
\end{algorithmic}
\end{algorithm}

\begin{algorithm}
\caption{Find the modal partition in $\Omega_{\Pi}$ for the classification problem.}
\label{alg:findNodePointerCat}
\begin{algorithmic}[1]
\Procedure{find\_modal\_partition}{$\mathcal{D}, prior\_arr$}
	\State $count_0\_arr \gets prior\_arr$
	\State $count_n\_arr \gets prior\_arr$
   	\For {each observation $j$ in $1,\dots,n$}\Comment{Adds category outcomes.}
   		\State $count_n\_arr[y_j] \gets count_n\_arr[y_j] + 1$
   	\EndFor
   	\State $p\_count \gets \text{sum}(count_0\_arr)$\Comment{Obtains pseudo count.}
   	\State $loglike_0 \gets \ln(B(count_0\_arr))$
   	\State $loglike_n \gets \ln(B(count_n\_arr))$
   	\State $loglike \gets loglike_n-loglike_0$
   	\State $logprob \gets loglike+\ln(p(\Pi_{0}))$
   	\State $\Pi^*\gets \text{create\_trivial\_part}(loglike, logprob)$
    \For {each dimension $r$ in $1,\dots,d$}
    	\State $m \gets 1$
    	\State $loglike_L \gets loglike_0$
    	\State $loglike_U \gets loglike_n$
    	\State $count_L\_arr \gets count_0\_arr$
    	\State $count_U\_arr \gets count_n\_arr$
    	\For {each observation $j$ in $1,\dots,n-1$}
	    	\State $loglike_L \gets loglike_L + \ln(count_L\_arr[y_{i_{j}}]/(j+p\_count - 1))$
	    	\State $count_L\_arr[y_{i_{j}^r}] \gets count_L\_arr[y_{i_{j}^r}] + 1$
	    	\State $count_U\_arr[y_{i_{j}^r}] \gets count_U\_arr[y_{i_{j}^r}] - 1$
    		\State $loglike_U \gets loglike_U - \ln(count_U\_arr[y_{i_{j}^r}]/(n+p\_count - j))$
        	\If {$x^r_{i_j^r} \neq x^r_{i_{j+1}^r}$}
        		\State $c \gets (x^r_{i_j^r} + x^r_{i_{j+1}^r})/2$
        		\State $loglike \gets loglike_L+loglike_U-2~loglike_0$
        		\State $logprob \gets loglike+\ln(p(\Pi_{r,m}))$
        		\State $m \gets m + 1$
        		\If {$\text{get\_logprob}(\Pi^*)<logprob$}
        			\State $\Pi^*\gets create\_part(c,r,loglike, logprob)$
        		\EndIf
        	\EndIf
		\EndFor
	\EndFor
	\State Return $\Pi^*$
\EndProcedure
\end{algorithmic}
\end{algorithm}

\begin{algorithm}
\caption{Fill the Bayesian greedy-modal tree and return the tree non-normalized log-probability.}
\label{alg:fill_tree}
\begin{algorithmic}[1]
\Procedure{tree\_logprob}{$\mathcal{D},root,\gamma$}
	\If {$root$ is a \emph{sprout}}
		\State Return $\text{fill\_tree}(\mathcal{D}, root, \gamma)$
   	\Else
   		\State $loglike \gets \text{get\_loglike}(\mathcal{D},\gamma)$\Comment{Applies equation \eqref{eq:likelihood}.}
   		\State $logprob \gets loglike+\ln(p(\Pi_{0}))$\Comment{Obtains $\ln(\widetilde{p})$.}
   		\State Return $logprob$
   	\EndIf
\EndProcedure
\item[]
\Procedure{fill\_tree}{$\mathcal{D}, sprout, \gamma$}
	\State $\mathcal{D}_{L}, \mathcal{D}_{U}\gets \text{split}(\mathcal{D}, sprout)$
	\State $\Pi_L \gets \text{choose\_partition}(\mathcal{D}_{L}, \gamma)$\Comment{Contains the search logic.}
    \State $child_L \gets \text{create\_node}(\Pi_L)$\Comment{Creates the node based on $\Pi_L$.}
    \State $\text{set\_lower\_child}(node, child_L)$
   	\State $\Pi_U \gets \text{choose\_partition}(\mathcal{D}_{U}, \gamma)$\Comment{Contains the search logic.}
    \State $child_U \gets \text{create\_node}(\Pi_U)$\Comment{Creates the node based on $\Pi_U$.}
    \State $\text{set\_upper\_child}(node, child_U)$
    \If {$child_L$ is a \emph{sprout}}
		\State $logprob_L \gets \text{fill\_tree}(\mathcal{D}_L, child_L, \gamma)$
    \Else
    	\State $logprob_L \gets \text{get\_logprob}(\Pi_L)$
	\EndIf
	\If {$child_U$ is a \emph{sprout}}
		\State $logprob_U \gets \text{fill\_tree}(\mathcal{D}_U, child_U, \gamma)$
    \Else
    	\State $logprob_U \gets \text{get\_logprob}(\Pi_U)$
	\EndIf
	\State Return $logprob_L + logprob_U$
\EndProcedure
\end{algorithmic}
\end{algorithm}

\end{document}